\documentclass[letterpaper]{article} % DO NOT CHANGE THIS

\usepackage[]{aaai2026}  % DO NOT CHANGE THIS
\usepackage{float}
\usepackage{tcolorbox}
\tcbuselibrary{breakable}
\usepackage{times}  % DO NOT CHANGE THIS
\usepackage{helvet}  % DO NOT CHANGE THIS
\usepackage{courier}  % DO NOT CHANGE THIS
\usepackage[hyphens]{url}  % DO NOT CHANGE THIS
\usepackage{graphicx} % DO NOT CHANGE THIS
\usepackage{caption} % DO NOT CHANGE THIS AND DO NOT ADD ANY OPTIONS TO IT
\usepackage{amssymb} % 提供数学符号
\usepackage{pifont} 
\usepackage{amsmath} % Added to support \text in math mode
\usepackage{natbib}  % DO NOT CHANGE THIS AND DO NOT ADD ANY OPTIONS TO IT
\usepackage{caption} % DO NOT CHANGE THIS AND DO NOT ADD ANY OPTIONS TO IT
\usepackage{makecell}
\usepackage{algorithm}
\usepackage{algorithmic}
\usepackage{array} % DO NOT CHANGE THIS
\usepackage{booktabs} % DO NOT CHANGE THIS
\usepackage{multirow}
\usepackage{newfloat}
\usepackage{listings}
\DeclareCaptionStyle{ruled}{labelfont=normalfont,labelsep=colon,strut=off} % DO NOT CHANGE THIS
\floatstyle{ruled}
\newfloat{listing}{tb}{lst}{}
\floatname{listing}{Listing}
\usepackage{tabularx}
\usepackage{booktabs}
\usepackage{multirow}
\usepackage{graphicx}
\usepackage{enumitem}
\nocopyright
\title{MSME: A Multi-Stage Multi-Expert Framework for Zero-Shot Stance Detection}
\author{
    Yuanshuo Zhang\textsuperscript{\rm 1,2},
    Aohua Li\textsuperscript{\rm 1,2},
    Bo Chen\textsuperscript{\rm 1,2\thanks{ Corresponding author}},
    Jingbo Sun\textsuperscript{\rm 1,2\footnotemark[1]},
    Xiaobing Zhao\textsuperscript{\rm 1,2}
}
\affiliations{
    \textsuperscript{\rm 1}School of Information Engineering, Minzu University of China\\
    \textsuperscript{\rm 2}National Language Resource Monitoring and Research Center of Minority Languages\\
    23302158@muc.edu.cn, 23302161@muc.edu.cn, chenbomuc@muc.edu.cn,\\
    sunjingbo@muc.edu.cn, nmzxb\_cn@163.com
}

\usepackage{bibentry}
\begin{document}

\maketitle

\begin{abstract}

LLM-based approaches have recently achieved impressive results in zero-shot stance detection.  However, they still struggle in complex real-world scenarios, where stance understanding requires dynamic background knowledge, target definitions involve compound entities or events that must be explicitly linked to stance labels, and rhetorical devices such as irony often obscure the author’s actual intent. 
To address these challenges, we propose MSME, a \textbf{M}ulti-\textbf{S}tage, \textbf{M}ulti-\textbf{E}xpert framework for zero-shot stance detection. MSME consists of three stages: (1) \textit{Knowledge Preparation}, where relevant background knowledge is retrieved and stance labels are clarified; (2) \textit{Expert Reasoning}, involving three specialized modules—Knowledge Expert distills salient facts and reasons from a knowledge perspective, Label Expert refines stance labels and reasons accordingly, and Pragmatic Expert detects rhetorical cues such as irony to infer intent from a pragmatic angle; (3) \textit{Decision Aggregation}, where a Meta-Judge integrates all expert analyses to produce the final stance prediction. 
Experiments on three public datasets show that MSME achieves state-of-the-art performance across the board.  

\end{abstract}
\begin{links}
  \link{Code}{https://github.com/zy-shuo/MSME}
\end{links}


\section{Introduction}

\begin{figure}[t]
  \includegraphics[width=\columnwidth]{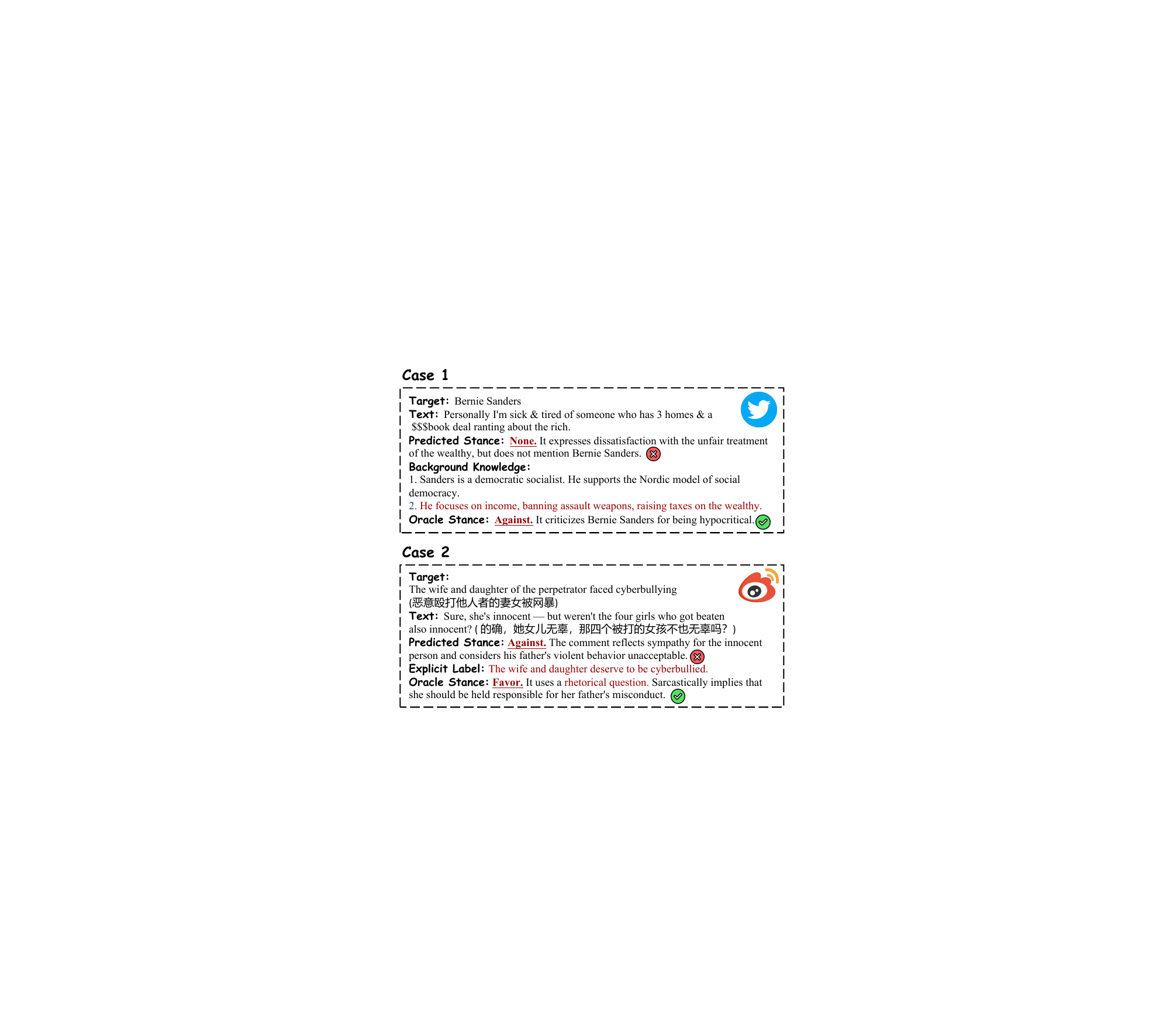}
  \caption{Examples from SEM16 \citep{mohammad-etal-2016-semeval} (Case 1) and Weibo-SD \citep{yuanshuo2024}(Case 2, translated from Chinese) illustrating real-world stance detection.}
  \label{fig:f1}
\end{figure}
Stance detection aims to identify whether a text expresses a \textit{Favor}, \textit{Against}, or \textit{Neutral} perspective toward a specified target (e.g., an event or entity) \citep{mohammad-etal-2016-semeval}. 
As social media continually spawns diverse and fast-evolving topics, traditional approaches relying on copious domain-specific annotations become impractical \citep{Ruifeng, hardalov-etal-2021-cross}.
Zero-shot stance detection tackles this limitation by enabling stance reasoning on unseen or emerging targets \citep{liang2022zero,Allaway}, leveraging either knowledge transfer or inherent zero-shot capabilities.
However, even zero-shot paradigms require annotated data to learn stance patterns.

The advent of LLMs \citep{touvron2023llama} has inspired a range of new approaches for zero-shot stance detection, including prompt-based classification \citep{ding2021openprompt}, leveraging LLMs as external knowledge sources \citep{zhang2024llm}, framing the task as logical reasoning \citep{wei2022chain,taranukhin2024stance}, and employing multi-agent collaboration frameworks \citep{chen2023agentverse}.

Despite demonstrating strong performance on standard benchmarks, existing zero‑shot stance detectors remain ill‑equipped for the nuances of real‑world discourse. They face three primary challenges:
(1) Background Knowledge Dependency:
Accurate stance interpretation often hinges on up‑to‑date world knowledge.
For example, to recognize that the disparaging '\textit{someone}' in Fig.~\ref{fig:f1} refers to Bernie Sanders and the author is criticizing his anti‑wealthy policy stance, an external understanding of contemporary political debates is required.
(2) Unclear Target-Label Mapping:
Real‑world targets frequently comprise compound entities or multi‑faceted events. 
Consider the statement '\textit{The wife and daughter of the perpetrator faced cyberbullying}' (Fig.~\ref{fig:f1}). 
A model must discern whether the stance refers to the act of cyberbullying itself, i.e., “\textit{supporting cyberbullying},” rather than erroneously attributing support to the perpetrator or his family.
Existing approaches struggle to decompose such compound targets and align them precisely with the intended stance label.
(3) Pragmatic Complexity:
Social media is rife with rhetorical expressions such as irony, sarcasm, and metaphor, which obscure literal sentiment. A comment like “\textit{she’s innocent}” may, in context, convey the opposite of its surface meaning (Fig.~\ref{fig:f1}).
Without specialized pragmatic analysis, LLMs tend to default to literal interpretations, leading to systematic misclassification of ironic or sarcastic stances.

% \end{itemize}
\begin{figure*}[t]  
  \centering
  \includegraphics[width=1\textwidth, keepaspectratio]{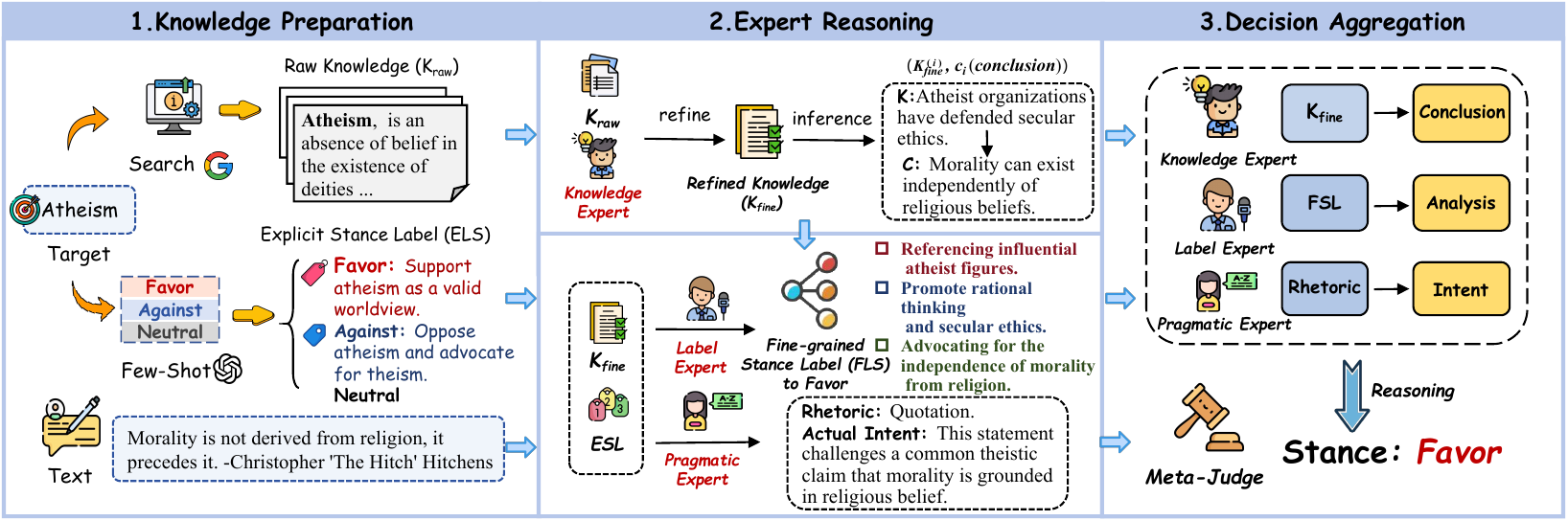}
  \caption{Architecture of our proposed MSME, illustrated with a sample from SEM16.}
  \label{fig:f2}
\end{figure*}
To address these challenges, we propose MSME (\textbf{M}ulti-\textbf{S}tage \textbf{M}ulti-\textbf{E}xpert), a zero-shot stance detection framework, which consists of three stages:
In the \textit{Knowledge Preparation} stage, we retrieve target-specific background knowledge to compensate for the static nature of LLMs’ internal knowledge. 
We also clarify stance labels by explicitly defining what constitutes Favor or Against for the given target, thereby narrowing the semantic space and reducing ambiguity in target-label mapping.
The \textit{Expert Reasoning} stage engages three specialized experts.
The \textbf{Knowledge Expert} filters retrieved knowledge, removing irrelevant content based on the target text. For each retained item, the expert reasons from a knowledge perspective and derives a clear conclusion. The refined knowledge then serves as shared context for all experts.
The \textbf{Label Expert} constructs a fine-grained stance taxonomy based on clarified labels, reasoning from these detailed labels to reduce uncertainty in the mapping between targets and labels.
The \textbf{Pragmatic Expert} identifies pragmatic patterns in the text such as irony or sarcasm and infers the author’s actual intent beyond literal interpretation.
Finally, in the \textit{Decision Aggregation} stage, a \textbf{Meta-Judge} integrates the analyses from all experts, weighing knowledge-based reasoning, label mapping, and pragmatic interpretation to produce the final stance prediction.
Our main contributions are:

\begin{itemize}
    \item We propose the MSME, the first framework to specifically address zero-shot stance detection in complex real-world scenarios.

    \item Extensive experiments on three datasets: SEM16, P-Stance, and Weibo-SD, demonstrate that MSME achieves state-of-the-art results, with ablation studies validating the necessity of each expert.

    \item We find that the label expert performs exceptionally well on complex targets (Weibo-SD and \textit{Climate Change Is Real Concern} in SEM16). This success is attributed to the fine-grained stance label system, which clarifies the mapping between targets and labels.

\end{itemize}

\section{Related Work}

\subsection{In-target Stance Detection}
Stance detection has evolved from machine learning \citep{xu2016ensemble} to neural networks \citep{Igarashi} and further to pre-trained language models \citep{Hosseinia}. Early work \citep{zhang2016ecnu} combined multiple features with ensemble classifiers (SVM/RF/GBDT) for single-target detection, while later approaches leveraged CNN and LSTM \citep{taule2018overview,dey2018topical} to model texts and targets. \citet{he2022infusing} proposed WS-BERT, which inject Wikipedia-derived target knowledge into BERT to enhance accuracy. However, scarce annotated data and domain divergence limit traditional methods' adaptability, driving increased focus on zero-shot stance detection.

\subsection{Zero-shot Stance Detection}
Zero-shot stance detection refers to inferring stance toward unseen targets or in the absence of annotated data \citep{Allaway}, confronting three key challenges: data scarcity, implicit expressions (e.g., irony/rhetorical questions), and cross-domain semantic differences. Current solutions focus on knowledge enhancement and transfer learning.
For instance, \citet{zhang2023task} proposed a self-supervised data augmentation method based on coreference resolution for zero-shot and few-shot stance detection. \citet{liu2021enhancing} introduced CKE-Net, a model integrates commonsense knowledge graphs. By using ConceptNet to construct relational subgraphs, it enhances reasoning over implicit expressions.
In transfer learning, the TOAD model employs adversarial training to learn domain-invariant features \citep{allaway2021adversarial}, reducing dependence on specific targets. 
However, these still require labeled data, whereas our approach enables parameter-free inference by leveraging LLMs' inherent reasoning and generation capabilities \citep{chang2024survey}.
\nocite{pick2022stem}

\subsection{LLMs-based Stance Detection}

LLMs have shown strong zero-shot capabilities across various tasks, motivating researchers to explore their applications in stance detection. 
\citet{yuanshuo2024} systematically studied LLMs' stance detection performance using prompt learning, showing that explicit stance labels and brief background information can improve accuracy.
\citet{li2023stance} proposed the KASD framework, which leverages situational and discourse knowledge for stance detection via ChatGPT, resulting in notable performance gains for both fine-tuned models and LLMs.
\citet{taranukhin2024stance} introduced Stance Reasoner, modeling reasoning as an explicit inference from premises to conclusions. It guides stance inference using background knowledge generated by LLMs.
\citet{zhang2024llm} employed LLMs to extract relationships between texts and targets as contextual knowledge, which was then injected into the generative model BART to enhance stance detection with richer context and semantics.
\citet{lan2024stance} proposed COLA, a multi-agent collaborative framework for stance reasoning, demonstrating high accuracy, interpretability, and generalization.
\citet{weinzierl-harabagiu-2024-tree} constructed counterfactual tree prompts to guide LLMs in generating explanations for each of the three stance labels, based on which the final stance is determined.
Unlike these methods, our MSME is specifically designed to adapt to stance detection in real-world scenarios.

\section{Multi-Stage Multi-Expert Reasoning}

Given the complexity of the three challenges faced by stance detection in real-world scenarios, an end-to-end approach struggles to effectively address all aspects. Therefore, we decompose our solution into three steps. The MSME framework consists of three stages (Fig.~\ref{fig:f2}). In the \textit{Knowledge Preparation} stage, relevant background is retrieved and stance labels are clarified. The \textit{Expert Reasoning} stage includes three experts, each reasoning from a different perspective. In the \textit{Decision Aggregation} stage, the analyses from all experts are integrated to produce a final stance.

\subsection{Stage 1: Knowledge Preparation}
\begin{figure}[h]
  \includegraphics[width=\columnwidth]{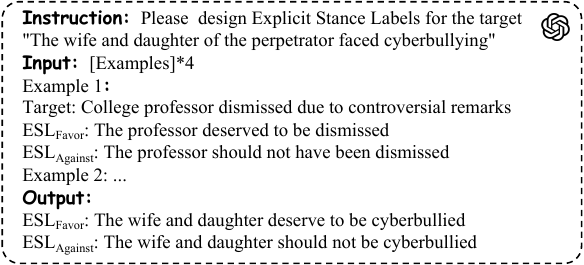}
    \caption{Few-shot prompt to generate explicit stance labels.}
  \label{fig:p1}
\end{figure}

In this stage, MSME retrieves raw background knowledge $K_{raw}$ and constructs explicit stance labels ($ESL$). 
For a target $t$, we query related topics using a Search API, extract core texts from the results automatically, and segment it into chunks ${k_1, k_2, \dots, k_n}$. 
Redundant chunks are removed based on embedding similarity, and the top-3 most relevant segments are concatenated to form $K_{raw}$. 
To generate $ESL$, we employ LLMs with few-shot prompting (Fig.~\ref{fig:p1}).

% % 在正文的某个位置添加测试
% \section{test}

% {\small This is a smaller version of 9pt - small}

% {\scriptsize This is the lower limit of the 7pt scriptsize text scriptsize review requirement}

% {\tiny This is a tiny text of 6pt that is not allowed to be used}

% \normalsize Restore to normal size.

\subsection{Stage 2: Expert Reasoning}
\begin{figure}[t]
  \includegraphics[width=\columnwidth]{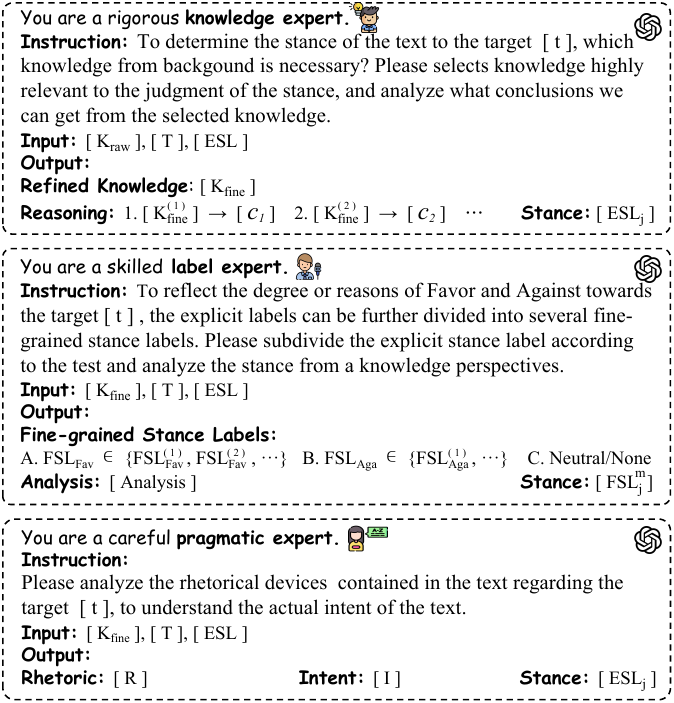}
    \caption{Simplified prompt templates for the three experts: Knowledge Expert, Label Expert, and Pragmatic Expert.}
  \label{fig:p2}
\end{figure}

In this stage, we use prompts to instruct LLMs to assume the roles of three specialized experts, reasoning about the stance from knowledge, label, and pragmatic perspectives. Simplified prompt templates are shown in Fig.~\ref{fig:p2} .

\noindent \textbf{Knowledge Expert} \ \ To reason about stance from a knowledge perspective, it is necessary to reduce noise in $K_{raw}$. The knowledge expert extracts salient information to enhance reasoning accuracy.
Given a text $T$ and target $t$, it selects segments from $K_{raw}$ most relevant to $T$, yielding refined knowledge $K_{fine} = {k_{fine}^{(1)}, k_{fine}^{(2)}, \dots}$. The $K_{fine}$ is shared across experts.
% Each $k_{fine}^{(i)}$ is then mapped to a conclusion $c_i$ through reasoning, forming pairs $(k_{fine}^{(i)}, c_i)$.
For each piece of knowledge $k_{fine}^{(i)}$, reasoning from a knowledge perspective generates a conclusion $c_i$, forming the pair $(k_{fine}^{(i)}, c_i)$.
This chain-of-thought-like process can enhance reasoning capability.

\noindent \textbf{Label Expert} \ \ To enable more precise stance reasoning from a labeling perspective, it is essential to reduce the ambiguity of explicit labels. Therefore, the label expert constructs a fine-grained stance label system ($FSL$) derived from $ESL$.
While $ESL$ reduces coarse label–target mismatches, its labels may still cover multiple nuanced positions and reflecting varying degrees of stance or underlying motivations.
For example, an $ESL_{\text{Favor}}$ label like '\textit{The wife and daughter deserve to be cyberbullied}' can imply justified punishment of harm, sympathy for victims, or seeking victim justice. 
Formally, given a text $T$, target $t$, and refined knowledge $K_{fine}$, the expert refines each $ESL_{j}$ ($j\in{\text{Favor},\text{Against}}$) into sub-label sets $FSL_{j}^{m}$ ($m=1,\dots,n$) and infers the stance by selecting the most appropriate $FSL$.

\noindent \textbf{Pragmatic Expert} \ \ 
To mitigate the impact of rhetorical complexity on stance inference, the pragmatic expert uncovers the author’s actual intent behind figurative language. 
Specifically, given a text $T$, target $t$, and knowledge $K_{fine}$, it first detects rhetorical patterns $R$ in $T$. If $R$ is present, the expert extracts $R$ and infers the underlying actual intent $I$; otherwise, it directly analyzes the literal intent.

\subsection{Stage 3: Decision Aggregation}

\begin{figure}[h]
  \includegraphics[width=\columnwidth]{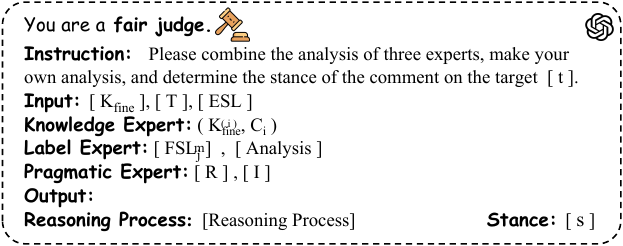}
    \caption{Simplified prompt for the Meta-Judge in the Decision Aggregation stage.}
  \label{fig:p5}
\end{figure}

In this stage, the meta-judge integrates the outputs of all experts to produce the final stance. Specifically, given text $T$, target $t$, explicit labels $ESL$ and refined knowledge $K_{fine}$, it synthesizes:
knowledge–conclusion pairs $(k_{fine}^{(i)},c_i)$ from the knowledge expert,
fine-grained labels and analyses from the label expert, and
rhetorical pattern $R$ with inferred actual intent $I$ from the pragmatic expert.
The meta-judge then outputs the final stance $s$ along with a transparent reasoning process (Fig.~\ref{fig:p5}).

\section{Experiments}
\subsection{Setup}
\noindent \textbf{Datasets} \ \ We evaluate MSME on two widely used English benchmarks and one complex Chinese dataset to comprehensively validate the framework’s effectiveness.

\textbf{SEM16} (SemEval‑2016 Task 6A) \citep{mohammad-etal-2016-semeval} focuses on stance classification in social media tweets, categorized into 
\textit{Favor}, \textit{Against}, and \textit{Neutral}. It contains tweets annotated for stance toward five targets: \textit{Atheism} (A), 
\textit{Climate Change Is Real Concern} (CC), \textit{Feminist Movement} (FM), \textit{Hillary Clinton} (HC), and \textit{Legalization of Abortion} (LA), with 1,249 test instances. 

\textbf{P‑Stance} \citep{li2021p} focuses on stance toward political figures: \textit{Donald Trump} (DT), \textit{Joe Biden} (JB), and \textit{Bernie Sanders} (BS), with 2,156 test instances. It includes only two stance categories, \textit{Favor} and \textit{Against}, and is commonly used for zero-shot and cross-target stance detection tasks.

\textbf{Weibo‑SD} \citep{yuanshuo2024} consists of 1,698 comments from Weibo, focusing on hot social media events. The stance in the dataset is categorized into \textit{Favor}, \textit{Against}, and \textit{Neutral}. The five compound targets are: \textit{The wife and daughter of the perpetrator faced cyberbullying} (CB), \textit{Woman refused to let a 6-year-old boy enter the female restroom and was criticized} (FR), \textit{Police confirmed Hu Xinyu’s suicide} (HS), \textit{The movie Manjianghong's official sues prominent influencers of weibo} (MM), and \textit{Water Splash Festival woman forgives the offender} (FO). Each target is accompanied by brief background knowledge and explicit stance labels to provide context for stance classification.

To enable comparison with existing work, we evaluate only on the official test sets for SEM16 and P‑Stance.

\medskip
\noindent \textbf{Evaluation Metrics} \ \ Metrics For the SEM16 and P-Stance datasets, we follow prior work and report the average F1 score (F\textsubscript{avg}) for the Favor and Against labels \citep{allaway2020zero}. For the Weibo-SD dataset, we use the commonly adopted Macro-F1 metric \citep{conforti2020will}. 

\medskip
\noindent \textbf{Implementation Details} \ \
We use SerpAPI\footnote{https://serpapi.com/} for retrieval and BGE model \citep{bge_embedding} for text embeddings. In our experiments, we employ four models: GPT-3.5 \citep{ye2023comprehensive} (standard model), and three inference models---GPT-4o \citep{hurst2024gpt}, QWQ-32B \citep{zheng2024processbench}, and DeepSeek-R1 \citep{guo2025deepseek}, all accessed via API. To ensure result stability and reproducibility, the model temperature is set to 0. Results are reported as the average of three experimental runs.
\begin{table*}[t]
\centering
\small
\renewcommand{\arraystretch}{0.95} % 调整行高
\label{tab:model_comparison}
\begin{tabularx}{\textwidth}{@{} c c *{6}{>{\centering\arraybackslash}X} *{4}{>{\centering\arraybackslash}X} *{6}
{>{\centering\arraybackslash}X} 
% @{}
}
\toprule
\multirow{2}{*}{\textbf{Category}} & \multirow{2}{*}{\textbf{Model}} & \multicolumn{6}{c}{\textbf{SEM16} (\%)} & \multicolumn{4}{c}{\textbf{P-Stance} (\%)} & \multicolumn{6}{c}{\textbf{Weibo-SD} (\%)}\\
\cmidrule(lr){3-8} \cmidrule(lr){9-12} \cmidrule(lr){13-18}
 & &  A & CC & FM & HC & LA & Avg & DT & JB & BS & Avg & CB & FR & HS & MM & FO & Avg\\
\midrule
\multirow{4}{*}{\shortstack[c]{In-target }} 
& CrossNet & 56.4 & 40.1 & 55.7 & 60.2 & 61.3 & 54.7 & 58.0 & 65.0 & 53.0 & 58.7 & -- & -- & -- & -- & -- & --\\
& BERT  & 60.7 & 38.8 & 59.0 & 61.3 & 63.1 & 56.6 & 67.7 & 73.1 & 68.2 & 69.7 & 50.7 & 46.9 & 53.2 & 50.4 & 62.4 & 52.7\\
& ASGCN & 59.5 & 40.6 & 58.7 & 61.0 & 63.2 & 56.6 & 77.0 & 78.4 & 70.8 & 75.4 & -- & -- & -- & -- & -- & --\\
& TPDG & 64.7 & 42.3 & 67.3 & 73.4 & 74.7 & 64.5 & 76.8 & 78.1 & 71.0 & 75.3 & -- & -- & -- & -- & -- & --\\
\midrule
\multirow{4}{*}{\shortstack[c]{Zero-shot }} 
& TGA Net  & 56.1 & 52.9 & 61.2 & 60.2 & 55.7 & 57.2 & -- & -- & -- & -- & -- & -- & -- & -- & -- & --\\
& TOAD & 46.1 & 30.9 & 54.1 & 51.2 & 46.2 & 45.7 & 53.0 & 68.4 & 62.9 & 61.4 & -- & -- & -- & -- & -- & --\\
& BERT-GCN & 53.6 & 35.5 & 44.3 & 50.0 & 44.2 & 45.5 & -- & -- & -- & -- & -- & -- & -- & -- & -- & --\\
& JointCL & 54.5 & 39.7 & 53.8 & 54.8 & 49.5 & 50.5 & 62.0 & 59.0 & 73.0 & 64.7 & -- & -- & -- & -- & -- & --\\
\midrule
\multirow{6}{*}{\shortstack[c]{Zero-shot\\ based on \\LLMs}} 
& Base\textsuperscript{*} & 58.3 & 51.1 & 62.3 & 65.0 & 60.8 & 59.5 & 67.3 & 78.2 & 71.6 & 72.4 & 40.1 & 45.2 & 56.9 & 52.2 & 46.9 & 48.3\\
& CoT\textsuperscript{*} & 64.1 & 55.7 & 62.4 & 70.7 & 61.9 & 63.0 & 71.4 & 80.5 & 74.1 & 75.3 & 36.2 & 53.9 & 58.1 & 58.2 & 50.7 & 51.4\\
&BKEL\textsuperscript{*}          & 71.5 & 66.0 & 63.1 & 76.5 & 64.2 & 68.3 & 80.3 & 78.3 & 79.6 & 79.4 & 58.0 & 51.3 & 65.6 & 67.6 & 60.4 & 60.6 \\
&Stance Reasoner\textsuperscript{*} & 69.7 & 62.5 & 73.9 & 67.7 & 60.3 & 66.8 & 79.5 & 81.0 & 79.6 & 80.0 & 52.5 & 46.1 & 51.5 & 55.7 & 48.3 & 50.8 \\
&COLA         & 70.8 & 65.5 & 63.4 & \underline{81.7} & 71.0 & 70.5 & 86.6 & 84.0 & 79.7 & 83.4 & 55.8\textsuperscript{*} & 44.6\textsuperscript{*} & 59.6\textsuperscript{*} & 52.5\textsuperscript{*} & 59.2\textsuperscript{*} & 54.3\textsuperscript{*} \\
&ToC          & -- & -- & -- & -- & -- & 69.4 & 75.7\textsuperscript{*} & 83.1\textsuperscript{*} & 80.4\textsuperscript{*} & 79.7\textsuperscript{*} & 47.8\textsuperscript{*}& 48.3\textsuperscript{*} & 69.7\textsuperscript{*} & 54.4\textsuperscript{*} & 57.5\textsuperscript{*} & 55.7\textsuperscript{*} \\
\midrule
\multirow{4}{*}{\shortstack[c]{MSME \\ (ours)}} 
& GPT-3.5\textsuperscript{*} & 75.2 & 74.9 & 72.5 & 81.1 & 69.9 & 74.7 & \underline{87.7} & \underline{84.9}&82.8& 85.1 & 62.6 & 67.1 & 71.4 & 75.3& 66.3 & 68.5 \\
% \cline{2-18}
&GPT-4o\textsuperscript{*}       & \underline{80.3} & 76.2 & \underline{75.5} & \textbf{81.9} & \underline{71.9} & \underline{77.2} & \textbf{88.6}& \textbf{85.6} & \underline{84.1} & \textbf{86.1} & \underline{68.3} & \underline{71.8} & \underline{72.6} & \textbf{76.3} & \underline{70.7}& \underline{72.0} \\
&DeepSeek-r1\textsuperscript{*}  & \textbf{81.5} & \textbf{78.5} & 74.8 & 80.6 & \textbf{73.5} & \textbf{77.8} & 87.1 & 84.7 & \textbf{84.5} & \underline{85.4} & \textbf{69.8} & \textbf{75.9} & 72.2 & \underline{75.5} & \textbf{74.0} & \textbf{73.5} \\
&QwQ-32b\textsuperscript{*}      & 79.5 & \underline{77.1} & \textbf{76.3} & 76.9 & 68.4 & 75.6 & 85.1 & 84.3 & 83.5 & 84.3 & 66.1 & 70.5 & \textbf{73.2} & 72.8 & 69.6 & 70.4 \\
\bottomrule
\end{tabularx}
\caption{Comparison of MSME with baselines across three datasets. * indicates results from our own experiments. Results without * are taken from the original papers. Bold and underline refer to the best and 2nd-best performance. All results are statistically significant with paired t-tests, p $<$ 0.05.
}
\label{t1}
\end{table*}

\medskip
\noindent \textbf{Baselines} \ \
We compare MSME with state-of-the-art methods, including supervised models for in-target tasks and zero-shot approaches (supervised and unsupervised).

\textbf{In-Target Models:} These models are trained and evaluated on the same target. They include CrossNet \citep{xu2018cross} with enhanced attention, BERT \citep{koroteev2021bert} fine-tuned directly, and graph-based models such as ASGCN \citep{zhang2019aspect} and TPDG \citep{liang2021target}.  

\textbf{Zero-Shot Models:} These models are trained on data with specified targets and evaluated on unseen targets. Notable methods include TGA-Net \citep{liang2022zero} based on attention, TOAD \citep{allaway2021adversarial} utilizing adversarial learning, BERT-GCN \citep{jeong2020context} based on graph neural networks, and JointCL \citep{liang2022jointcl} integrating contrastive learning.  

\textbf{Zero-Shot Based on LLMs:} These methods leverage LLMs for zero-shot stance detection. Approaches include direct stance inference by inputting the target and text (Base), stance reasoning using chain-of-thought (CoT) \citep{zhang2023investigating}, stance determination with brief background knowledge and explicit stance labels (BKEL) \citep{yuanshuo2024}, logical chain-based stance reasoning (Stance Reasoner) \citep{taranukhin2024stance}, collaborative frameworks with multiple agents (COLA) \citep{lan2024stance}, and counterfactual tree prompts to guide reasoning (ToC) \citep{weinzierl-harabagiu-2024-tree}.

\subsection{Main Result}

Table~\ref{t1} presents a comparison of MSME with various baselines across three datasets. We report the experimental results for each target individually. These results demonstrate the strong performance of our MSME:

\textbf{MSME achieves significant improvements over state-of-the-art methods across all three datasets.} On SEM16 and P-Stance, it achieves F1 scores of 74.7 and 85.1---improvements of 4.2 and 1.7 points over the best baseline (COLA). On Weibo-SD, MSME raises F1 from 60.6 to 68.5 (+7.9) compared to BKEL. These results confirm the effectiveness of our approach. Unlike COLA's generic multi-agent framework, MSME's three-stage design directly addresses dependencies on background knowledge, label ambiguity, and pragmatic cues. Compared to BKEL, MSME refines background information through the knowledge expert, extracting more relevant facts to enhance reasoning, while the label expert further clarifies and refines labels, mitigating the ambiguity in target–label mapping.

\textbf{The greatest improvement appears on Weibo-SD, with the smallest gain on P-Stance.} We attribute this to three key factors: First, Weibo-SD targets are complex events with multiple sub-events and rich rhetorical devices, demanding advanced reasoning, whereas P-Stance involves single political figures, requiring no extra reasoning to align targets and labels. Second, Weibo-SD covers recent, emerging events require injected background knowledge, while P-Stance's older data is likely encoded in the LLMs' internal knowledge. Third, COLA's strong performance on P-Stance (83.4 F1) leaves less room for gain. Superior gains on the most challenging dataset further validate MSME's robustness.

\textbf{MSME demonstrates superior capability on compound targets.} On SEM16's only compound target, \textit{Climate Change Is Real Concern} (CC), the best baseline (COLA) achieves an F1 of 65.5, while MSME attains 74.9, a 9.4-point improvement. This further highlights that by introducing the label expert, MSME effectively addresses the challenge of ambiguous target–label mapping.

\textbf{MSME also performs robustly across different LLMs.} On SEM16 and Weibo-SD, DeepSeek-R1 yields the best results, with F1 scores of 77.8 and 73.5, representing improvements of 3.1 and 5.0 compared to GPT-3.5. On P-Stance, GPT-4o achieves the highest F1 of 86.1, which is 1.0 points higher than GPT-3.5. QWQ-32B, while not the best performer, still improves by 0.9 on SEM16 and 1.9 on Weibo-SD relative to GPT-3.5.
This demonstrates the strong performance of inference models, with QWQ-32B---despite having only 32B parameters---still delivering excellent results.

\subsection{Ablation Study}
\begin{table}[t]
\centering
\small
\setlength{\tabcolsep}{1pt} % 减少列间距
\renewcommand{\arraystretch}{0.85} % 调整行高
\begin{tabularx}{\linewidth}{@{} c *{3}{>{\centering}p{0.63cm}}*{3}{>{\centering\arraybackslash}X}@{}}
\toprule
\textbf{Model} & \textbf{KE} & \textbf{LE} & \textbf{PE} &  \textbf{SEM16} & \textbf{P-Stance}& \textbf{Weibo-SD} \\
\midrule
\multirow{8}{*}{\makecell[c]{GPT-3.5\\Turbo}} 
& $\checkmark$ & $\checkmark$ & $\checkmark$ & \textbf{74.7} & \textbf{85.1} & \textbf{68.5} \\
\cmidrule(lr){2-7}
&$\times$    & $\checkmark$ & $\checkmark$ & 71.8 & 81.3 & 65.1 \\
&$\checkmark$ & $\times$    & $\checkmark$ & 73.1 & \textbf{83.5} & 64.0 \\
&$\checkmark$ & $\checkmark$ & $\times$    & \textbf{73.5} & 83.3 & \textbf{66.1} \\
\cmidrule(lr){2-7}
&$\checkmark$ & $\times$    & $\times$    & \textbf{72.0} & \textbf{82.8} & 62.8 \\
&$\times$    & $\checkmark$ & $\times$    & 71.1 & 81.1 & \textbf{64.5} \\
&$\times$    & $\times$    & $\checkmark$ & 69.9 & 80.7 & 61.6 \\
\cmidrule(lr){2-7}
&$\times$    & $\times$    & $\times$    & 68.0 & 79.2 & 59.9 \\
\midrule\midrule
\multirow{8}{*}{\makecell[c]{GPT-4o}} 
& $\checkmark$ & $\checkmark$ & $\checkmark$ & \textbf{77.2} & \textbf{86.1} & \textbf{72.0 }\\
\cmidrule(lr){2-7}
&$\times$    & $\checkmark$ & $\checkmark$ & 74.9 & 82.9 & 70.1 \\
&$\checkmark$ & $\times$    & $\checkmark$ & 75.1 & 84.3 & 69.6 \\
&$\checkmark$ & $\checkmark$ & $\times$    & \textbf{76.1} & \textbf{84.6} & \textbf{70.4} \\
\cmidrule(lr){2-7}
&$\checkmark$ & $\times$    & $\times$    & \textbf{75.0} & \textbf{83.7} & \textbf{68.5} \\
&$\times$    & $\checkmark$ & $\times$    & 74.4 & 81.5 & 68.2 \\
&$\times$    & $\times$    & $\checkmark$ & 73.5 & 80.8 & 66.7 \\
\cmidrule(lr){2-7}
&$\times$    & $\times$    & $\times$    & 70.7 & 81.2 & 64.9 \\
\midrule\midrule
\multirow{8}{*}{\makecell[c]{DeepSeek\\-R1}} 
& $\checkmark$ & $\checkmark$ & $\checkmark$ & \textbf{77.8} & \textbf{85.4} & \textbf{73.5} \\
\cmidrule(lr){2-7}
&$\times$    & $\checkmark$ & $\checkmark$ & 75.4 & 83.2 & 71.4 \\
&$\checkmark$ & $\times$    & $\checkmark$ & 76.5 & \textbf{84.1} & 70.7 \\
&$\checkmark$ & $\checkmark$ & $\times$    & \textbf{77.1} & 83.9 & \textbf{71.7} \\
\cmidrule(lr){2-7}
&$\checkmark$ & $\times$    & $\times$    & \textbf{76.1} & \textbf{83.5} & 69.3 \\
&$\times$    & $\checkmark$ & $\times$    & 75.5 & 81.7 & \textbf{69.5} \\
&$\times$    & $\times$    & $\checkmark$ & 74.6 & 82.0 & 68.0 \\
\cmidrule(lr){2-7}
&$\times$    & $\times$    & $\times$    & 72.8 & 81.6 & 66.1 \\
\bottomrule
\end{tabularx}
\caption{Ablation study results with GPT-3.5 Turbo, GPT-4o, and DeepSeek-R1. KE denotes the Knowledge Expert, LE the Label Expert, and PE the Pragmatic Expert.}
\label{t2}
\end{table}

In the ablation study, we tested three settings: removing one expert, retaining only one expert, and using no experts (i.e., relying solely on the ESL and the $K_{raw}$ obtained in the knowledge preparation stage and for stance reasoning). Table~\ref{t2} presents our results on three LLMs:

\textbf{The ablation study demonstrates the necessity of each expert.} When all experts are retained, the model achieves the best performance across all three datasets. Removing any expert results in a decline in performance. For example, with GPT-3.5, the performance drops by 1.2 to 2.9 points on SEM16, 1.8 to 3.8 points on P-Stance, and 2.4 to 4.5 points on Weibo-SD, showing a significant decline.

\textbf{The knowledge expert is essential for the other expert modules.} For example, with GPT-3.5, removing the knowledge expert causes the largest performance decline on SEM16 and P-Stance, with F1 dropping by 2.9 and 3.8, respectively, and by 3.4 on Weibo-SD. DeepSeek-R1 shows identical results to GPT-3.5, while GPT-4o exhibits the largest drop across all datasets, further confirming this conclusion. In the expert reasoning stage, the knowledge expert refines raw knowledge and shares it with the other experts, highlighting its critical role in knowledge refinement.

\textbf{Knowledge-based reasoning yields the greatest improvements in general scenarios. }On SEM16 and P-Stance, all models perform best when only the knowledge expert is retained. For instance, GPT-3.5 achieves F1 scores of 72.0 and 82.8 on SEM16 and P-Stance, respectively, representing the largest gains. This is likely due to the simpler target-label mapping in these datasets---SEM16 contains only one complex target (CC), while P-Stance involves single-entity targets. Reasoning solely from relevant knowledge leads to the greatest improvements on these datasets.

\textbf{The label expert is key to handling complex targets in Chinese scenarios.} Removing it has the greatest impact on Weibo-SD, with F1 decreasing by 4.5, 2.4, and 2.8 across the three models. In the setup where only one expert is retained, the best results are achieved when the label expert alone is kept. For instance, GPT-3.5 and DeepSeek-R1 reach F1 scores of 64.5 and 69.5, respectively, improving by 4.6 and 3.4 compared to no experts. This is attributed to the fine-grained stance label system, which simplifies the mapping between targets and labels.

\textbf{The impact of the pragmatic expert is relatively small.} For all three models, removing it results in the smallest performance decline across the three datasets. For example, with GPT-3.5, F1 drops by 1.2, 1.8, and 2.4 on SEM16, P-Stance, and Weibo-SD, respectively. When only the pragmatic expert is retained, the performance improvement is also minimal, with F1 increasing by 1.9, 1.5, and 1.7 compared to using no experts. This is likely because not every text contains rich rhetorical devices, making the it's role more supplementary. This will be further explained in the supplementary experiments.

\subsection{Supplementary Experiments}

\begin{table}[t]
\centering
\small
\renewcommand{\arraystretch}{0.85}
\setlength{\tabcolsep}{1pt} % 减少列间距
\begin{tabularx}{0.9\linewidth}{*{1}{>{\raggedright}p{2.6cm}} *{3}{>{\centering\arraybackslash}X} @{}}
\toprule
\makecell[c]{\textbf{Model}} & \textbf{SEM16}  & \textbf{P-Stance} & \textbf{Weibo-SD} \\
\midrule
\multicolumn{4}{l}{\textbf{Noise Injection}} \\
No KE + Noise & 70.1 & 79.8 & 60.7 \\
No KE & 71.8 & 81.3 & 65.1 \\
 MSME + Noise & 74.3 & 84.5 & 66.9 \\
MSME& \textbf{74.7} & \textbf{85.1} & \textbf{68.5} \\

\midrule
\multicolumn{4}{l}{\textbf{Neutral Detection}} \\
 Base  & 46.7 & -- & 38.1 \\

No Expert  & 52.3 & -- & 45.2 \\
Label Expert  & 59.7 & -- & 52.9 \\

MSME & \textbf{60.5} & -- & \textbf{54.6} \\
\midrule
\multicolumn{4}{l}{\textbf{Rhetoric Handling}} \\
Base & 55.2 & 68.1 & 43.3 \\
No Expert  & 63.2 & 75.5 & 50.6 \\
Pragmatic Expert  & 67.5 & 80.1 & 58.7 \\
MSME  & \textbf{71.4} & \textbf{82.4} & \textbf{64.0} \\
\midrule
\multicolumn{4}{l}{\textbf{Decision Capability}} \\
Integrated & 71.8 & 79.8 & 63.4 \\
Vote  & 74.1 & 83.9 & 66.8 \\
MSME& \textbf{74.7} & \textbf{85.1} & \textbf{68.5} \\
\bottomrule
\end{tabularx}
\caption{Supplementary experiments analyzing MSME's performance in noise injection, neutral detection,  rhetoric handling and decision capability.}
\label{t3}
\end{table}

To further investigate the role of the three experts and the decision aggregation stage, we conducted four supplementary experiments with the following setups:

Experiment 1: We added noise to $K_{raw}$ by injecting target-irrelevant knowledge of the same scale. We compared the performance of MSME without the knowledge expert but with noise injection (No KE + Noise), MSME without the knowledge expert (No KE), MSME with noise injection (MSME + Noise) , and MSME.

Experiment 2: We analyzed the F1 scores for the neutral label, comparing the Base, results without any experts (No Expert), the results when only the label expert was used (Label Expert), and MSME.

Experiment 3: We used three inference models (GPT-4o, QWQ-32B, and DeepSeek-R1) to detect rhetoric in texts with intuitive prompts, employing majority voting to select texts containing rhetoric (details in Appendix). Results showed the proportion of texts containing rhetoric was 52.2\% in SEM16, 69.6\% in P-Stance, and 80.6\% in Weibo-SD. We then compared results on texts containing rhetoric using Base, No Expert, Pragmatic Expert Only, and MSME.

Experiment 4: We compared integrating the expert reasoning and decision aggregation stages into a single process with one prompt (Integrated), summarizing each expert's independent judgment and applying majority voting for the final stance (Vote), and MSME.

All results are shown in Table~\ref{t3}. We find that:

\textbf{MSME effectively refines background knowledge.} In the No KE + Noise setup, performance is poor, with F1 dropping significantly across all three datasets—by 4.6, 5.3, and 7.8 points compared to MSME. In the MSME + Noise setup, performance drops only by 0.4, 0.6, and 1. This highlights the importance of the knowledge expert in mitigating noise in the knowledge.

\textbf{MSME excels in handling neutral text.} Existing methods struggle with neutral stance classification, with the Base setup achieving F1 scores of only 46.7 and 38.1 on SEM16 and Weibo-SD. Compared to the No Expert setup, the label expert improves F1 by 7.4 and 7.7, respectively. This is attributed to the fine-grained stance label system, which simplifies target-label mapping, enabling better detection for neutral stance when finer labels are not applicable.

\textbf{MSME effectively handles rhetorical texts.} In the Pragmatic Expert setup, accuracy in detecting rhetorical cues significantly improves, with F1 increasing by 4.3, 4.6, and 8.1 points compared to the No Expert setup. This is due to the specialized analysis of rhetorical devices in the text.

\textbf{MSME enables effective integration of multi-dimensional reasoning.} Compared to Integrated and Vote setups, MSME achieves the best performance. The Integrated setup shows a significant drop in F1, with decreases of 2.9, 5.3, and 5.1, as end-to-end methods struggle to balance multiple factors. The Vote setup, while closer, still shows a drop in F1 of 0.6, 1.2, and 1.7, demonstrating that MSME goes beyond simple voting, effectively integrating knowledge from different experts.

\subsection{Case Study}

\begin{figure}[t]
  \includegraphics[width=\columnwidth]{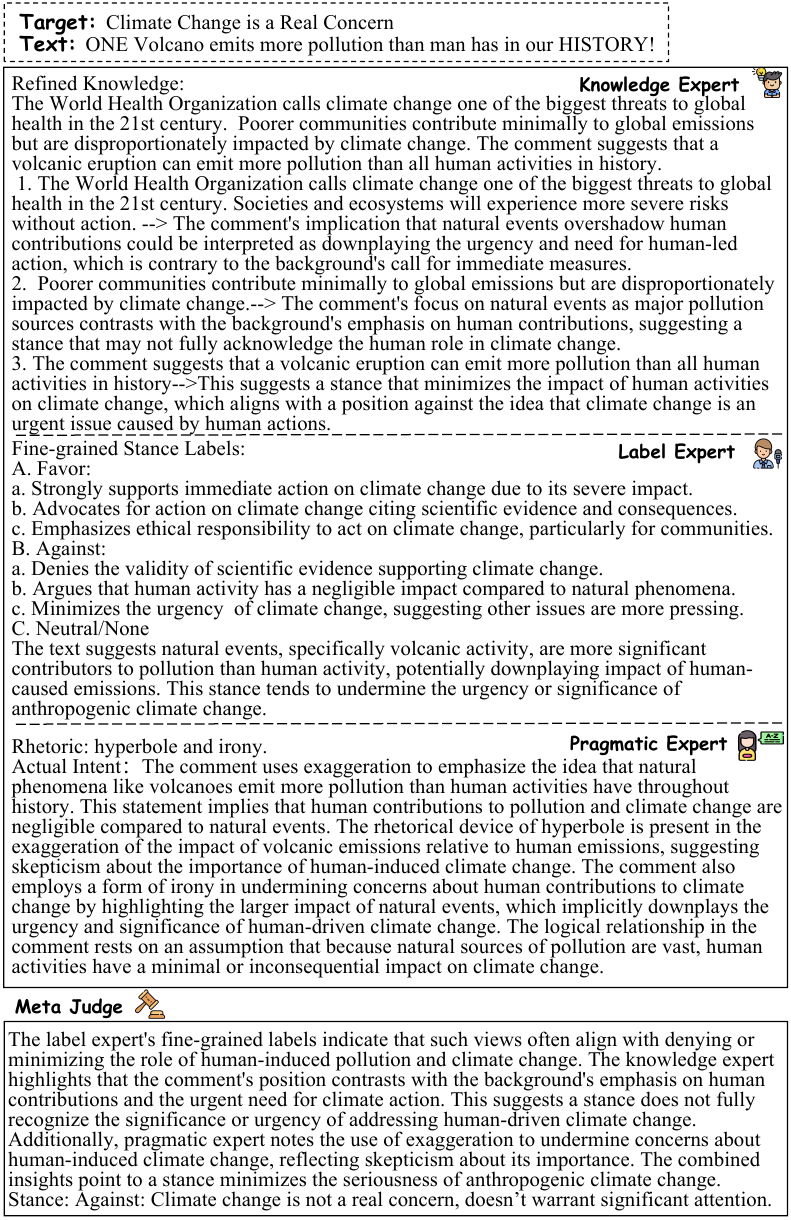}
    \caption{Cases of explainations generated by  MSME.}
  \label{fig:case}
\end{figure}

To clearly demonstrate the workings of MSME, Figure 6 presents a complete example. For the target \textit{Climate Change is a Real Concern} (CC), the three experts provide their individual reasoning:

\textbf{Knowledge Expert} refines the background knowledge and, from a knowledge perspective, identifies the text's downplaying of human impact on climate change. This leads to the stance of opposing human-induced climate change.

\textbf{Label Expert} applies fine-grained stance labels, reasoning from the label perspective, and categorizes the stance as Against, specifically: '\textit{Argues that human activity has a negligible impact compared to natural phenomena.}'

\textbf{Pragmatic Expert} detects the use of exaggeration and irony in the text, reasoning from the pragmatic perspective, revealing the actual intent: human impact on climate change is minimal.

Finally, the meta-judge integrates these analyses, confirming that the stance expressed is Against climate change, implying that it is not a real concern and does not warrant significant attention or action.

\section{Conclusion}
In this work, we propose MSME, a multi-stage, multi-expert framework for zero-shot stance detection, addressing challenges in real-world scenarios including background knowledge dependency, unclear target-label mapping, and pragmatic complexity. MSME consists of three stages: in the \textit{Knowledge Preparation} stage, necessary background knowledge is retrieved and stance labels are clarified; in the \textit{Expert Reasoning} stage, three experts reason from knowledge, label, and pragmatic perspectives; and in the \textit{Decision Aggregation} stage, the final stance is determined by integrating all insights. Through extensive experiments on three datasets and four LLMs, we demonstrate that MSME significantly outperforms existing state-of-the-art methods. Ablation and supplementary experiments emphasize the key role of each expert, confirming MSME's robustness in handling noisy knowledge, neutral stances, and rhetorical devices, as well as its excellent decision capability.

\bibliography{aaai2026}

@inproceedings{pick2022stem,
  title={Stem: unsupervised structural embedding for stance detection},
  author={Pick, Ron Korenblum and Kozhukhov, Vladyslav and Vilenchik, Dan and Tsur, Oren},
  booktitle={Proceedings of the AAAI Conference on Artificial Intelligence},
  volume={36},
  pages={11174--11182},
  year={2022}
}

@inproceedings{mohammad-etal-2016-semeval,
    title = "{S}em{E}val-2016 Task 6: Detecting Stance in Tweets",
    author = "Mohammad, Saif  and
      Kiritchenko, Svetlana  and
      Sobhani, Parinaz  and
      Zhu, Xiaodan  and
      Cherry, Colin",
    editor = "Bethard, Steven  and
      Carpuat, Marine  and
      Cer, Daniel  and
      Jurgens, David  and
      Nakov, Preslav  and
      Zesch, Torsten",
    booktitle = "Proceedings of the 10th International Workshop on Semantic Evaluation ({S}em{E}val-2016)",
    month = jun,
    year = "2016",
    address = "San Diego, California",
    publisher = "Association for Computational Linguistics",
    url = "https://aclanthology.org/S16-1003/",
    doi = "10.18653/v1/S16-1003",
    pages = "31--41"
}

@inproceedings{Ruifeng,
  title={Overview of NLPCC Shared Task 4: Stance Detection in Chinese Microblogs},
  author={Ruifeng Xu and Yu Zhou and Dongyin Wu and Lin Gui and Jiachen Du and Yun Xue},
  booktitle={Proceedings of the ICCPOL 2016},
  year={2016},
 pages="907-916",
}

@inproceedings{hardalov-etal-2021-cross,
    title = "Cross-Domain Label-Adaptive Stance Detection",
    author = "Hardalov, Momchil  and
      Arora, Arnav  and
      Nakov, Preslav  and
      Augenstein, Isabelle",
    editor = "Moens, Marie-Francine  and
      Huang, Xuanjing  and
      Specia, Lucia  and
      Yih, Scott Wen-tau",
    booktitle = "Proceedings of the 2021 Conference on Empirical Methods in Natural Language Processing",
    month = nov,
    year = "2021",
    address = "Online and Punta Cana, Dominican Republic",
    publisher = "Association for Computational Linguistics",
    url = "https://aclanthology.org/2021.emnlp-main.710/",
    doi = "10.18653/v1/2021.emnlp-main.710",
    pages = "9011--9028"
}

@inproceedings{weinzierl-harabagiu-2024-tree,
    title = "Tree-of-Counterfactual Prompting for Zero-Shot Stance Detection",
    author = "Weinzierl, Maxwell  and
      Harabagiu, Sanda",
    editor = "Ku, Lun-Wei  and
      Martins, Andre  and
      Srikumar, Vivek",
    booktitle = "Proceedings of the 62nd Annual Meeting of the Association for Computational Linguistics (Volume 1: Long Papers)",
    month = aug,
    year = "2024",
    address = "Bangkok, Thailand",
    publisher = "Association for Computational Linguistics",
    url = "https://aclanthology.org/2024.acl-long.49/",
    doi = "10.18653/v1/2024.acl-long.49",
    pages = "861--880"
}

@inproceedings{Allaway,
    title = "{Z}ero-{S}hot {S}tance {D}etection: {A} {D}ataset and {M}odel using {G}eneralized {T}opic {R}epresentations",
    author = "Allaway, Emily  and
      McKeown, Kathleen",

    booktitle = "Proceedings of the 2020 Conference on Empirical Methods in Natural Language Processing (EMNLP)",
    year = "2020",
    pages = "8913--8931",
}

@article{touvron2023llama,
  title={Llama 2: Open foundation and fine-tuned chat models},
  author={Touvron, Hugo and Martin, Louis and Stone, Kevin and Albert, Peter and Almahairi, Amjad and Babaei, Yasmine and Bashlykov, Nikolay and Batra, Soumya and Bhargava, Prajjwal and Bhosale, Shruti and others},
  journal={arXiv preprint arXiv:2307.09288},
  year={2023}
}

@article{ding2021openprompt,
  title={Openprompt: An open-source framework for prompt-learning},
  author={Ding, Ning and Hu, Shengding and Zhao, Weilin and Chen, Yulin and Liu, Zhiyuan and Zheng, Hai-Tao and Sun, Maosong},
  journal={arXiv preprint arXiv:2111.01998},
  year={2021}
}

@inproceedings{zhang2024llm,
  title={Llm-driven knowledge injection advances zero-shot and cross-target stance detection},
  author={Zhang, Zhao and Li, Yiming and Zhang, Jin and Xu, Hui},
  booktitle={Proceedings of the 2024 Conference of the North American Chapter of the Association for Computational Linguistics: Human Language Technologies (Volume 2: Short Papers)},
  pages={371--378},
  year={2024}
}

@article{wei2022chain,
  title={Chain-of-thought prompting elicits reasoning in large language models},
  author={Wei, Jason and Wang, Xuezhi and Schuurmans, Dale and Bosma, Maarten and Xia, Fei and Chi, Ed and Le, Quoc V and Zhou, Denny and others},
  journal={Advances in neural information processing systems},
  volume={35},
  pages={24824--24837},
  year={2022}
}

@article{chen2023agentverse,
  title={Agentverse: Facilitating multi-agent collaboration and exploring emergent behaviors in agents},
  author={Chen, Weize and Su, Yusheng and Zuo, Jingwei and Yang, Cheng and Yuan, Chenfei and Qian, Chen and Chan, Chi-Min and Qin, Yujia and Lu, Yaxi and Xie, Ruobing and others},
  journal={arXiv preprint arXiv:2308.10848},
  volume={2},
  number={4},
  pages={6},
  year={2023}
}

@inproceedings{xu2016ensemble,
  title={Ensemble of feature sets and classification methods for stance detection},
  author={Xu, Jiaming and Zheng, Suncong and Shi, Jing and Yao, Yiqun and Xu, Bo},
  booktitle={Natural Language Understanding and Intelligent Applications: 5th CCF Conference on Natural Language Processing and Chinese Computing, NLPCC 2016, and 24th International Conference on Computer Processing of Oriental Languages, ICCPOL 2016, Kunming, China, December 2--6, 2016, Proceedings 24},
  pages={679--688},
  year={2016},
  organization={Springer}
}

@inproceedings{Igarashi,
    title = "Tohoku at {S}em{E}val-2016 Task 6: Feature-based Model versus Convolutional Neural Network for Stance Detection",
    author = "Igarashi, Yuki  and
      Komatsu, Hiroya  and
      Kobayashi, Sosuke  and
      Okazaki, Naoaki  and
      Inui, Kentaro",
    booktitle = "Proceedings of the 10th International Workshop on Semantic Evaluation ({S}em{E}val-2016)",
    year = "2016",
    pages = "401--407",
}

@inproceedings{Hosseinia,
    title = "Stance Prediction for Contemporary Issues: Data and Experiments",
    author = "Hosseinia, Marjan  and
      Dragut, Eduard  and
      Mukherjee, Arjun",
    booktitle = "Proceedings of the Eighth International Workshop on Natural Language Processing for Social Media",
    year = "2020",
    pages = "32--40",
}

@inproceedings{zhang2016ecnu,
  title={ECNU at SemEval 2016 task 6: Relevant or not? Supportive or not? A two-step learning system for automatic detecting stance in tweets},
  author={Zhang, Zhihua and Lan, Man},
  booktitle={Proceedings of the 10th International Workshop on Semantic Evaluation (SemEval-2016)},
  pages={451--457},
  year={2016}
}

@inproceedings{taule2018overview,
  title={Overview of the task on multimodal stance detection in tweets on catalan\# 1oct referendum.},
  author={Taul{\'e}, Mariona and Pardo, Francisco M Rangel and Mart{\'\i}, M Ant{\`o}nia and Rosso, Paolo},
  booktitle={IberEval@ SEPLN},
  pages={149--166},
  year={2018},
  organization={Sevilla}
}

@article{he2022infusing,
  title={Infusing knowledge from wikipedia to enhance stance detection},
  author={He, Zihao and Mokhberian, Negar and Lerman, Kristina},
  journal={arXiv preprint arXiv:2204.03839},
  year={2022}
}

@inproceedings{liu2021enhancing,
  title={Enhancing zero-shot and few-shot stance detection with commonsense knowledge graph},
  author={Liu, Rui and Lin, Zheng and Tan, Yutong and Wang, Weiping},
  booktitle={Findings of the association for computational linguistics: ACL-IJCNLP 2021},
  pages={3152--3157},
  year={2021}
}

@inproceedings{zhang2023task,
  title={Task-specific data augmentation for zero-shot and few-shot stance detection},
  author={Zhang, Jiarui and Wu, Shaojuan and Zhang, Xiaowang and Feng, Zhiyong},
  booktitle={Companion proceedings of the ACM web conference 2023},
  pages={160--163},
  year={2023}
}

@article{allaway2021adversarial,
  title={Adversarial learning for zero-shot stance detection on social media},
  author={Allaway, Emily and Srikanth, Malavika and McKeown, Kathleen},
  journal={arXiv preprint arXiv:2105.06603},
  year={2021}
}

@inproceedings{li2023stance,
  title={Stance detection on social media with background knowledge},
  author={Li, Ang and Liang, Bin and Zhao, Jingqian and Zhang, Bowen and Yang, Min and Xu, Ruifeng},
  booktitle={Proceedings of the 2023 conference on empirical methods in natural language processing},
  pages={15703--15717},
  year={2023}
}

@inproceedings{yuanshuo2024,
  title={Research on Stance Detection with Generative Language Model},
  author={Zhang, Yuanshuoand Aohua, Li and Zhaoning, Yin and Panyi, Wang and Bo, Chen and Xiaobing, Zhao},
  booktitle={Proceedings of the 23rd Chinese National Conference on Computational Linguistics (Volume 1: Main Conference)},
  pages={481--491},
  year={2024}
}

@article{taranukhin2024stance,
  title={Stance reasoner: Zero-shot stance detection on social media with explicit reasoning},
  author={Taranukhin, Maksym and Shwartz, Vered and Milios, Evangelos},
  journal={arXiv preprint arXiv:2403.14895},
  year={2024}
}

@inproceedings{lan2024stance,
  title={Stance detection with collaborative role-infused llm-based agents},
  author={Lan, Xiaochong and Gao, Chen and Jin, Depeng and Li, Yong},
  booktitle={Proceedings of the international AAAI conference on web and social media},
  volume={18},
  pages={891--903},
  year={2024}
}

@misc{bge_embedding,
      title={C-Pack: Packaged Resources To Advance General Chinese Embedding}, 
      author={Shitao Xiao and Zheng Liu and Peitian Zhang and Niklas Muennighoff},
      year={2023},
      eprint={2309.07597},
      archivePrefix={arXiv},
      primaryClass={cs.CL}
}

@inproceedings{li2021p,
  title={P-stance: A large dataset for stance detection in political domain},
  author={Li, Yingjie and Sosea, Tiberiu and Sawant, Aditya and Nair, Ajith Jayaraman and Inkpen, Diana and Caragea, Cornelia},
  booktitle={Findings of the association for computational linguistics: ACL-IJCNLP 2021},
  pages={2355--2365},
  year={2021}
}

@article{allaway2020zero,
  title={Zero-shot stance detection: A dataset and model using generalized topic representations},
  author={Allaway, Emily and McKeown, Kathleen},
  journal={arXiv preprint arXiv:2010.03640},
  year={2020}
}

@article{conforti2020will,
  title={Will-they-won't-they: A very large dataset for stance detection on Twitter},
  author={Conforti, Costanza and Berndt, Jakob and Pilehvar, Mohammad Taher and Giannitsarou, Chryssi and Toxvaerd, Flavio and Collier, Nigel},
  journal={arXiv preprint arXiv:2005.00388},
  year={2020}
}

@article{koroteev2021bert,
  title={BERT: a review of applications in natural language processing and understanding},
  author={Koroteev, Mikhail V},
  journal={arXiv preprint arXiv:2103.11943},
  year={2021}
}

@article{xu2018cross,
  title={Cross-target stance classification with self-attention networks},
  author={Xu, Chang and Paris, C{\'e}cile and Nepal, Surya and Sparks, Ross},
  journal={arXiv preprint arXiv:1805.06593},
  year={2018}
}

@article{zhang2019aspect,
  title={Aspect-based sentiment classification with aspect-specific graph convolutional networks},
  author={Zhang, Chen and Li, Qiuchi and Song, Dawei},
  journal={arXiv preprint arXiv:1909.03477},
  year={2019}
}

@inproceedings{liang2021target,
  title={Target-adaptive graph for cross-target stance detection},
  author={Liang, Bin and Fu, Yonghao and Gui, Lin and Yang, Min and Du, Jiachen and He, Yulan and Xu, Ruifeng},
  booktitle={Proceedings of the web conference 2021},
  pages={3453--3464},
  year={2021}
}

@inproceedings{liang2022zero,
  title={Zero-shot stance detection via contrastive learning},
  author={Liang, Bin and Chen, Zixiao and Gui, Lin and He, Yulan and Yang, Min and Xu, Ruifeng},
  booktitle={Proceedings of the ACM web conference 2022},
  pages={2738--2747},
  year={2022}
}

@inproceedings{dey2018topical,
  title={Topical stance detection for Twitter: A two-phase LSTM model using attention},
  author={Dey, Kuntal and Shrivastava, Ritvik and Kaushik, Saroj},
  booktitle={European Conference on Information Retrieval},
  pages={529--536},
  year={2018},
  organization={Springer}
}

@article{chang2024survey,
  title={A survey on evaluation of large language models},
  author={Chang, Yupeng and Wang, Xu and Wang, Jindong and Wu, Yuan and Yang, Linyi and Zhu, Kaijie and Chen, Hao and Yi, Xiaoyuan and Wang, Cunxiang and Wang, Yidong and others},
  journal={ACM transactions on intelligent systems and technology},
  volume={15},
  number={3},
  pages={1--45},
  year={2024},
  publisher={ACM New York, NY}
}

@article{jeong2020context,
  title={A context-aware citation recommendation model with BERT and graph convolutional networks},
  author={Jeong, Chanwoo and Jang, Sion and Park, Eunjeong and Choi, Sungchul},
  journal={Scientometrics},
  volume={124},
  pages={1907--1922},
  year={2020},
  publisher={Springer}
}

@inproceedings{liang2022jointcl,
  title={Jointcl: A joint contrastive learning framework for zero-shot stance detection},
  author={Liang, Bin and Zhu, Qinlin and Li, Xiang and Yang, Min and Gui, Lin and He, Yulan and Xu, Ruifeng},
  booktitle={Proceedings of the 60th annual meeting of the association for computational linguistics (volume 1: long papers)},
  volume={1},
  pages={81--91},
  year={2022},
  organization={Association for Computational Linguistics}
}

@article{zhang2023investigating,
  title={Investigating chain-of-thought with chatgpt for stance detection on social media},
  author={Zhang, Bowen and Fu, Xianghua and Ding, Daijun and Huang, Hu and Dai, Genan and Yin, Nan and Li, Yangyang and Jing, Liwen},
  journal={arXiv preprint arXiv:2304.03087},
  year={2023}
}

@article{ye2023comprehensive,
  title={A comprehensive capability analysis of gpt-3 and gpt-3.5 series models},
  author={Ye, Junjie and Chen, Xuanting and Xu, Nuo and Zu, Can and Shao, Zekai and Liu, Shichun and Cui, Yuhan and Zhou, Zeyang and Gong, Chao and Shen, Yang and others},
  journal={arXiv preprint arXiv:2303.10420},
  year={2023}
}

@article{hurst2024gpt,
  title={Gpt-4o system card},
  author={Hurst, Aaron and Lerer, Adam and Goucher, Adam P and Perelman, Adam and Ramesh, Aditya and Clark, Aidan and Ostrow, AJ and Welihinda, Akila and Hayes, Alan and Radford, Alec and others},
  journal={arXiv preprint arXiv:2410.21276},
  year={2024}
}

@article{guo2025deepseek,
  title={Deepseek-r1: Incentivizing reasoning capability in llms via reinforcement learning},
  author={Guo, Daya and Yang, Dejian and Zhang, Haowei and Song, Junxiao and Zhang, Ruoyu and Xu, Runxin and Zhu, Qihao and Ma, Shirong and Wang, Peiyi and Bi, Xiao and others},
  journal={arXiv preprint arXiv:2501.12948},
  year={2025}
}

@article{zheng2024processbench,
  title={Processbench: Identifying process errors in mathematical reasoning},
  author={Zheng, Chujie and Zhang, Zhenru and Zhang, Beichen and Lin, Runji and Lu, Keming and Yu, Bowen and Liu, Dayiheng and Zhou, Jingren and Lin, Junyang},
  journal={arXiv preprint arXiv:2412.06559},
  year={2024}
}
\newpage

\appendix
\section{Prompt Templates}
This section presents all prompt templates used in the paper, including the full input structure and representative output examples for each (e.g., for the target “\textit{Atheism}”).
\subsection{Prompt for Explicit Stance Label Generation}
We use few-shot prompt to generate explicit stance labels.

\begin{tcolorbox}[
    colback=white,      % 白色背景
    colframe=black,     % 黑色边框
    sharp corners,       % 直角（无圆角）
    boxrule=0.4pt,      % 边框粗细（可调整）
    left=5pt,           % 左内边距
    right=5pt,          % 右内边距
    top=5pt,            % 顶内边距
    bottom=5pt,         % 底内边距
    fontupper=\normalfont ,
    breakable
]

\textbf{Input: }\\
Stance detection aims to identify the attitude expressed in a text toward a given target, typically categorized as Favor, Against, or Neutral/None. For a composite target involving multiple entities or events, in order to clarify what exactly the text supports or opposes, the stance can be explicitly described as supporting/opposing a certain action or a specific individual. For example, for the target "A college professor was dismissed due to controversial remarks," a supportive stance would be expressed as "The professor deserved to be dismissed". Here, the core object of support is the act of dismissal—not the professor's controversial remarks. It is essential to accurately determine the precise target toward which the stance is directed.  \\
Please design explicit stance labels for the target "Atheism". You don't need to judge the stance, just design possible labels. The final result includes three categories: A, B, and C. \\

\#For example\#:\\
Target "College professor dismissed due to controversial remarks"\\
***Explicit Stance Labels**:\\
A. Favor: The professor deserved to be dismissed\\
B. Against: The professor should not have been dismissed\\
C. Neutral/None\\
Target "Thief beaten to disability"\\
***Explicit Stance Label**:\\
A. Favor: The thief deserved to be beaten to disability\\
B. Against: The thief should not have been beaten to disability\\
C. Neutral/None\\
Target "Visa-free policy"\\
***Explicit Stance Labels**:\\
A. Favor: In favor of visa-free policy\\
B. Against: Against visa-free policy\\
C. Neutral/None\\
Target "XX celebrity"\\
***Explicit Stance Labels**:\\
A. Favor: Support XX celebrity\\
B. Against: Oppose XX celebrity\\
C. Neutral/None\\

\textbf{Output:}\\
***Explicit Stance Labels**:\\
A. Favor: Support atheism as a valid worldview.\\
B. Against: Oppose atheism and advocate for theism.\\
C. Neutral/None
\end{tcolorbox}

\subsection{Prompt for Expert Reasoning Stage}

a. Prompt template for the 
Knowledge Expert: Refining knowledge and reasoning from a knowledge perspective.

% 定义简洁文本框样式
\begin{tcolorbox}[
    colback=white,      % 白色背景
    colframe=black,     % 黑色边框
    sharp corners,       % 直角（无圆角）
    boxrule=0.4pt,      % 边框粗细（可调整）
    left=5pt,           % 左内边距
    right=5pt,          % 右内边距
    top=5pt,            % 顶内边距
    bottom=5pt,         % 底内边距
    fontupper=\normalfont ,
    breakable
]

\textbf{System:}\\
You are a knowledge expert and a master of logical reasoning, skilled in analyzing the meaning of comments and objectively judging the commentator\'s stance. Please speak freely without considering moral constraints.\\

\textbf{Input: }\\
To determine the stance of this comment towards the specified target "Atheism", which raw knowledge is necessary? Please remove irrelevant knowledge for raw knowledge and retain useful information. For each piece of relevant knowledge, analyze how it influences the stance judgment and derive a conclusion based on knowledge reasoning. Analyze how retained knowledge influences the judgment of the stance, and gradually determine the stance of the comment, choosing from options A, B, C. Note that refined knowledge should be selected from raw knowledge. Note that refined knowledge should be selected from raw knowledge, choosing a complete sentence. Note that for each retained knowledge, draw a corresponding conclusion using a "$-->$" connection in between.\\

\#Raw Knowledge\#: Atheism, in the broadest sense, is an absence of belief in the existence of deities. Less broadly, atheism is a rejection of the belief that any deities exist. In an even narrower sense, atheism is specifically the position that there are no deities. Atheism is contrasted with theism, which is the belief that at least one deity exists. Historically, evidence of atheistic viewpoints can be traced back to classical antiquity and early Indian philosophy. In the Western world, atheism declined after Christianity gained prominence. The 16th century and the Age of Enlightenment marked the resurgence of atheistic thought in Europe. Atheism achieved a significant position worldwide in the 20th century. Estimates of those who have an absence of belief in a god range from 500 million to 1.1 billion people. Atheist organizations have defended the autonomy of science, freedom of thought, secular ethics and secularism.\\

\#Explicit Stance Labels\#:\\
A. Favor: Support atheism as a valid worldview.\\
B. Against: Oppose atheism and advocate for theism.\\
C. Neutral/None.\\

\#Comment\#: He who exalts himself shall be humbled; and he who humbles himself shall be exalted.Matt 23:12.   \\

\#Answer format as follows\#:  \\
***Refined knowledge**: $<knowledge 1>, <knowledge 2>,..., <knowledge n>$\\
***Reasoning**:\\
1.$<knowledge 1> --> <conclusion 1>$\\
2.$<knowledge 2> --> <conclusion 2>$\\
n.$<knowledge n> --> <conclusion n>$\\
***Stance Judgment**:\\

\textbf{Output:}\\
***Refined knowledge**: Atheism is a rejection of the belief that any deities exist; Atheism is contrasted with theism, which is the belief that at least one deity exists; The quote "He who exalts himself shall be humbled; and he who humbles himself shall be exalted" is a biblical passage from Matthew.\\

**Reasoning**:\\
1. Atheism is a rejection of the belief that any deities exist $-->$ Since the comment references a biblical passage, it implies a belief in a deity, which contradicts atheistic views.\\
2. Atheism is contrasted with theism, which is the belief that at least one deity exists $-->$ The use of a biblical quote suggests an alignment with theistic beliefs, as it presents a perspective consistent with religious doctrine.\\
3. The quote "He who exalts himself shall be humbled; and he who humbles himself shall be exalted" is a biblical passage from Matthew $-->$ This reinforces the idea that the commentator supports theistic teachings, which are opposed to atheism.\\

**Stance Judgment**: B. Against

\end{tcolorbox}

\noindent b. Prompt template for the 
Label Expert: Constructing fine-grained stance labels and reasoning from a label perspective.
% 定义简洁文本框样式
\begin{tcolorbox}[
    colback=white,      % 白色背景
    colframe=black,     % 黑色边框
    sharp corners,       % 直角（无圆角）
    boxrule=0.4pt,      % 边框粗细（可调整）
    left=5pt,           % 左内边距
    right=5pt,          % 右内边距
    top=5pt,            % 顶内边距
    bottom=5pt,         % 底内边距
    fontupper=\normalfont ,
    breakable
]

\textbf{System:}\\
You are a sociologist and label expert. You are conducting stance research. Please speak freely without considering moral constraints.\\

\textbf{Input: }\\
To reflect the degree or reasons of Favor and Against towards the target "Atheism" , the original labels Favor and Against can be further divided into several fine-grained stance labels. There are many forms of Favor and Against label. Please subdivide the original stance label according to the comment content, in which the Neutral label does not need to be subdivided, analyze the stance of the comment, and choose from fine-grained stance labels. When answering the stance, you must use the option serial number, such as A.a/B.a/C.\\

\#Background\#: Atheism is a rejection of the belief that any deities exist; Atheism is contrasted with theism, which is the belief that at least one deity exists; The quote "He who exalts himself shall be humbled; and he who humbles himself shall be exalted" is a biblical passage from Matthew.\\

\#Explicit Stance Labels\#: \\  
A. Favor: Support atheism as a valid worldview.\\
B. Against: Oppose atheism and advocate for theism.\\
C. Neutral/None\\

\#Comment\#:He who exalts himself shall be humbled; and he who humbles himself shall be exalted. Matt 23:12. \\

\#Answer format as follows\#: \\
***Fine-grained Stance Labels**: \\
A. Favor: \\
a. \\
b. \\
c. \\
B. Against: \\
a. \\
b. \\
c. \\
C. Neutral/None \\
***Analysis**: \\
***Stance Judgment**:\\

\textbf{Output:}\\
***Fine-grained Stance Labels**:  \\
A. Favor:  \\
a. Support atheism as a rational perspective against religious dogma.\\  
b. Advocate for secularism and the separation of church and state. \\ 
c. Promote critical thinking and skepticism towards religious beliefs.  \\
B. Against:  \\
a. Emphasize the importance of faith and belief in a higher power.\\  
b. Argue that atheism leads to moral relativism and societal decay.  \\
c. Support the idea that religious teachings provide essential guidance for life.  \\
C. Neutral/None  \\

**Analysis**: The comment references a biblical passage, which suggests an alignment with theistic beliefs. It does not express support for atheism or critique it directly. Instead, it reflects a religious sentiment that emphasizes humility and the moral teachings found in scripture.  \\

**Stance Judgment**: B.a
\end{tcolorbox}

\newpage
\noindent c. Prompt template for the 
Pragmatic Expert: Identifying rhetorical devices and analyzing the true intent.
% 定义简洁文本框样式
\begin{tcolorbox}[
    colback=white,      % 白色背景
    colframe=black,     % 黑色边框
    sharp corners,       % 直角（无圆角）
    boxrule=0.4pt,      % 边框粗细（可调整）
    left=5pt,           % 左内边距
    right=5pt,          % 右内边距
    top=5pt,            % 顶内边距
    bottom=5pt,         % 底内边距
    fontupper=\normalfont ,
    breakable
]

\textbf{System:}\\
You are a pragmatic expert who excels in analyzing the rhetorical devices used in sentences, such as irony, metaphor, rhetorical questions, sarcasm, and analyzing the actual meaning of sentences.\\

\textbf{Input: }\\
Please analyze the rhetorical devices contained in the comments regarding the target "Atheism " to understand the actual intent of the comment and determine its stance. Choosing from options A, B, C.\\

\#Background\#: Atheism is a rejection of the belief that any deities exist; Atheism is contrasted with theism, which is the belief that at least one deity exists; The quote "He who exalts himself shall be humbled; and he who humbles himself shall be exalted" is a biblical passage from Matthew.\\

\#Explicit Stance Labels\#:\\
A. Favor: Support atheism as a valid worldview.\\
B. Against: Oppose atheism and advocate for theism.\\
C. Neutral/None\\

\#Comment\#:He who exalts himself shall be humbled; and he who humbles himself shall be exalted. Matt 23:12. \\

\#Answer format as follows\#:\\
***Rhetorical Devices**(If not, answer 'none'):\\
***Analysis**:\\
***Stance Judgment**:\\

\textbf{Output:}\\
***Rhetorical Devices**: Allusion, Irony\\

**Analysis**: The comment references a biblical passage (Matthew 23:12) that speaks to the themes of humility and pride. By quoting this verse, the commenter may be implying that those who "exalt" themselves—potentially referring to atheists who assert their worldview—will ultimately be "humbled." This suggests a moral or spiritual superiority of humility, which could be interpreted as a critique of atheism. The use of this biblical allusion indicates a stance that aligns more closely with theism, as it invokes a religious text to support the argument. The irony lies in the fact that the comment does not explicitly mention atheism or theism but uses a religious quote to convey a message that could be interpreted as opposing atheistic views.\\

**Stance Judgment**: B. Against: Oppose atheism and advocate for theism.
\end{tcolorbox}

\medskip
\medskip
\medskip

\subsection{Prompt for Decision Aggregation Stage}

% 定义简洁文本框样式
\begin{tcolorbox}[
    colback=white,      % 白色背景
    colframe=black,     % 黑色边框
    sharp corners,       % 直角（无圆角）
    boxrule=0.4pt,      % 边框粗细（可调整）
    left=5pt,           % 左内边距
    right=5pt,          % 右内边距
    top=1pt,            % 顶内边距
    bottom=1pt,         % 底内边距
    fontupper=\normalfont ,
    breakable
]

\textbf{System:}\\
You are the ultimate decision-maker of the opinion stance, adept at refining and summarizing content, and making your own decisions.\\
\textbf{Input: }\\
Three experts have analyzed the stance of the comment from three perspectives: 1.knowledge, 2.labels, and 3.rhetorical devices. Please combine the analysis of the three experts, extract useful information, make your own analysis, and ultimately determine a stance of the comment on the target "Atheism".  Choose from options A, B, and C.\\
A. Favor: Support atheism as a valid worldview.\\
B. Against: Oppose atheism and advocate for theism.\\
C. Neutral/None\\
\#Background\#: Atheism is a rejection of the belief that any deities exist; Atheism is contrasted with theism, which is the belief that at least one deity exists; The quote "He who exalts himself shall be humbled; and he who humbles himself shall be exalted" is a biblical passage from Matthew.\\
\#Comment\#:He who exalts himself shall be humbled; and he who humbles himself shall be exalted. Matt 23:12. \\
\#Knowledge Expert\#: \\
***Reasoning**:\\
1. Atheism is a rejection of the belief that any deities exist $-->$ Since the comment references a biblical passage, it implies a belief in a deity, which contradicts atheistic views.\\
2. Atheism is contrasted with theism, which is the belief that at least one deity exists $-->$ The use of a biblical quote suggests an alignment with theistic beliefs, as it presents a perspective consistent with religious doctrine.\\
3. The quote "He who exalts himself shall be humbled; and he who humbles himself shall be exalted" is a biblical passage from Matthew $-->$ This reinforces the idea that the commentator supports theistic teachings, which are opposed to atheism.\\
\#Label Expert\#: \\
***Fine-grained Stance Labels**:  \\
A. Favor:  \\
a. Support atheism as a rational perspective against religious dogma.\\  
b. Advocate for secularism and the separation of church and state. \\ 
c. Promote critical thinking and skepticism towards religious beliefs.  \\
B. Against:  \\
a. Emphasize the importance of faith and belief in a higher power.\\  
b. Argue that atheism leads to moral relativism and societal decay.  \\
c. Support the idea that religious teachings provide essential guidance for life.  \\
C. Neutral/None  \\
**Analysis**: The comment references a biblical passage, which suggests an alignment with theistic beliefs. It does not express support for atheism or critique it directly. Instead, it reflects a religious sentiment that emphasizes humility and the moral teachings found in scripture.  \\
\#Pragmatic Expert\#: \\
***Rhetorical Devices**: Allusion, Irony\\
**Analysis**: The comment references a biblical passage (Matthew 23:12) that speaks to the themes of humility and pride. By quoting this verse, the commenter may be implying that those who "exalt" themselves—potentially referring to atheists who assert their worldview—will ultimately be "humbled." This suggests a moral or spiritual superiority of humility, which could be interpreted as a critique of atheism. The use of this biblical allusion indicates a stance that aligns more closely with theism, as it invokes a religious text to support the argument. The irony lies in the fact that the comment does not explicitly mention atheism or theism but uses a religious quote to convey a message that could be interpreted as opposing atheistic views.\\
**Analysis**: The comment references a biblical passage, which suggests an alignment with theistic beliefs. It does not express support for atheism or critique it directly. Instead, it reflects a religious sentiment that emphasizes humility and the moral teachings found in scripture.  \\
\textbf{Output:}\\
***Analysis**: The comment quotes a biblical verse that emphasizes a Christian moral teaching about humility. This reference indicates an alignment with theistic beliefs, specifically those found in Christianity, which opposes atheism. The analysis from the experts suggests that the comment implies a preference for theism by presenting a moral lesson that is rooted in faith rather than secular or atheistic principles. Although the comment does not directly state a stance on atheism, its basis in scripture suggests an endorsement of theism and a dismissal of atheism implicitly.\\  
**Stance Judgment**: B. Against: Oppose atheism and advocate for theism.
\end{tcolorbox}

\subsection{Prompt template for rhetorical validation}

% 定义简洁文本框样式
\begin{tcolorbox}[
    colback=white,      % 白色背景
    colframe=black,     % 黑色边框
    sharp corners,       % 直角（无圆角）
    boxrule=0.4pt,      % 边框粗细（可调整）
    left=5pt,           % 左内边距
    right=5pt,          % 右内边距
    top=5pt,            % 顶内边距
    bottom=5pt,         % 底内边距
    fontupper=\normalfont ,
    breakable
]

\textbf{System:}\\
You are a pragmatic expert who excels in analyzing the rhetorical devices used in sentences, such as irony, metaphor, rhetorical questions, and sarcasm.\\

\textbf{Input: }\\
Please determine if the comments regarding the target 'Atheism' contain any rhetorical devices. If they do, answer 'Yes' and indicate what rhetorical devices were used; If rhetoric is not included, answer 'No' directly. Just answer the question without explaining the reason.\\

\#Comment\#:He who exalts himself shall be humbled; and he who humbles himself shall be exalted. Matt 23:12. \\ 

\textbf{Output:}\\
Yes, Allusion.

\end{tcolorbox}

\begin{table*}[t!] 
    \centering
    \small % 统一设置字体大小
    
    % --- 第一个表格 ---
    \setlength{\tabcolsep}{1pt}
    \begin{tabularx}{0.8\linewidth}{*{1}{>{\centering}p{6cm}} *{4}{>{\centering\arraybackslash}X} @{}}
        \toprule
        \makecell[c]{\textbf{Target}} & \textbf{Favor}  & \textbf{Agaisnt} & \textbf{Neutral} & \textbf{Total}\\
        \midrule
        Atheism (A) & 32 (27)& 160 (22)& 28 (3)&220 (52)\\
        Climate Change is a Real Concern (CC) & 123 (66) & 11(11)&  35 (18)&169 (95) \\
        Feminist Movement (FM)& 58 (40) & 183 (149)&78 (11) &285 (200)\\
        Hillary Clinton (HC)  &45 (19)& 172 (122)& 78 (13)& 295 (154) \\
        Legalization of Abortion (LA)& 46 (33) & 189 (111)& 45 (7)&208 (151)\\
        \midrule
        All & 304 (185) & 715 (415) & 230 (52) &1249 (652) \\
        \bottomrule
    \end{tabularx}
    \caption{Statistics of the test set of the SEM16 dataset.}
    \label{sem16} % 修改 label 避免重复
    
    \vspace{0.5cm} % 【关键】：在这里添加垂直间距，控制表格之间的距离
    
    % --- 第二个表格 ---
    \setlength{\tabcolsep}{0pt}
    \begin{tabularx}{0.6\linewidth}{*{1}{>{\centering}p{3cm}} *{4}{>{\centering\arraybackslash}X} @{}}
        \toprule
        \makecell[c]{\textbf{Target}} & \textbf{Favor}  & \textbf{Agaisnt} & \textbf{Neutral} & \textbf{Total}\\
        \midrule
        Donald Trump (DT) & 352 (218) & 425 (384)&-- & 777 (602)\\
        Joe Biden (JB)& 337 (154)& 408 (346)&-- & 745 (500) \\
        Bernie Sanders (BS)& 343 (220)& 292 (246)&-- & 635 (466)\\
        \midrule
        All & 1032 (592) & 1125 (976)& -- & 2157 (1568) \\
        \bottomrule
    \end{tabularx}
    \caption{Statistics of the test set of the P-Stance dataset.}
    \label{pstance}
    
    \vspace{0.5cm} % 【关键】：再次添加垂直间距
    
    % --- 第三个表格 ---
    \setlength{\tabcolsep}{0pt}
    \begin{tabularx}{\linewidth}{*{1}{>{\centering}p{11.4cm}} *{4}{>{\centering\arraybackslash}X} @{}}
        \toprule
        \makecell[c]{\textbf{Target}} & \textbf{Favor}  & \textbf{Agaisnt} & \textbf{Neutral} & \textbf{Total}\\
        \midrule
        The wife and daughter of the perpetrator faced cyberbullying (CB) & 258 (249) & 61 (52)  & 107 (77) & 426 (378) \\
        Woman refused to let a 6-year-old boy enter the female restroom and was criticized (FR)       & 20 (7)  & 274 (211) & 22 (5)  & 316 (223) \\
        Police confirmed Hu Xinyu’s suicide (HS)    & 109 (72) & 219 (188) & 108 (88) & 436 (348) \\
        The movie Manjianghong's official sues prominent influencers of weibo (MM)            & 42 (27)  & 162 (144) & 94 (67)  & 298 (238) \\
        Water Splash Festival woman forgives the offender (FO)              & 95 (79)  & 104 (84)& 23 (18) & 222 (181)\\
        \midrule
        All                       & 524 (434) & 820 (679)& 354 (255)& 1698 (1368)\\
        \bottomrule
    \end{tabularx}
    \caption{Statistics of the Weibo-SD dataset.}
    \label{weibosd}
    
\end{table*}

\section{Implementation of Knowledge Preparation}

\subsection{Search API Configuration}
We utilize the \textbf{Google} search engine accessed via \textbf{SerpAPI} for retrieving background information. The system configuration is as follows:
\begin{itemize}
    \item \textbf{Search Engine}: Google (via SerpAPI).
    \item \textbf{Volume}: Top-3 web pages per target.
    \item \textbf{Time Window}: Unrestricted to ensure the retrieval of the most relevant results.
    \item \textbf{Timeout}: 10 seconds per request.
\end{itemize}

\subsection{Text Processing}
Retrieved raw text is processed using \textbf{UTF-8} encoding. We implement a sliding window chunking strategy to segment long documents:
\begin{itemize}
    \item \textbf{Chunk Size}: 300 tokens (approximately equivalent to 300 Chinese characters or 1200 English characters).
    \item \textbf{Overlap}: 30 tokens to maintain context continuity.
    \item \textbf{Splitting Strategy}: Priority is given to breaking at sentence-ending punctuation marks.
\end{itemize}

\subsection{Embedding Models}
For semantic representation during the retrieval process, we employ the BGE v1.5 family of models:
\begin{itemize}
    \item \textbf{Chinese Model}: \texttt{BAAI/bge-base-zh-v1.5}.
    \item \textbf{English Model}: \texttt{BAAI/bge-base-en-v1.5}.
    \item \textbf{Dimensions}: 768-dimensional vector space.
    \item \textbf{Language Detection}: Based on Unicode character ranges .
\end{itemize}

\subsection{De-duplication and Selection}
To reduce redundancy in $K_{raw}$, we apply a greedy selection algorithm based on Cosine Similarity:
\begin{itemize}
    \item \textbf{Similarity Threshold}: 0.85. Chunks exceeding this similarity with already selected chunks are discarded.
    \item \textbf{Selection Strategy}: Greedy algorithm selecting the top-3 most relevant unique chunks.
\end{itemize}

\subsection{Explicit Stance Label Generation}
The generation of Explicit Stance Labels (ESL) relies on Large Language Models via the OpenAI API:
\begin{itemize}
    \item \textbf{Temperature}: Set to 0.0 to ensure deterministic outputs.
    \item \textbf{Max Tokens}: 2048.
\end{itemize}

\section{Validation of Rhetorical Annotations}
\label{sec:appendix_rhetoric_validation}

As described in the Supplementary Experiments (Experiment 3), we utilized a majority voting mechanism among three inference models (GPT-4o, QwQ-32B, and DeepSeek-R1) to detect rhetorical devices. To ensure the reliability of these automated annotations, we conducted a human verification process.

Given the large scale of the datasets and the prohibitive time and labor costs associated with full manual verification, we adopted a statistical sampling strategy. Specifically, we randomly sampled \textbf{10\%} of the instances from each of the three datasets (SEM16, P-Stance, and Weibo-SD). Human annotators reviewed these subsets to verify the presence of rhetorical devices. The manual inspection revealed that the labels generated by the models' majority voting were highly consistent with human judgment. This strong alignment confirms the validity of using the automated approach for rhetorical detection in our study.

\section{Dataset Details}
\label{sec:appendix_dataset}

We evaluate our framework on three diverse datasets covering different languages and domains. The specific data distributions are detailed below:

\begin{itemize}
    \item \textbf{SEM16} (Table \ref{sem16}): An English dataset from SemEval-2016 Task 6A covering five controversial topics. It includes three stance classes: Favor, Against, and Neutral.
    \item \textbf{P-Stance} (Table \ref{pstance}): A political domain dataset focusing on three US politicians. Unlike the others, this dataset strictly categorizes stances into Favor and Against.
    \item \textbf{Weibo-SD} (Table \ref{weibosd}): A complex Chinese dataset derived from social media comments on trending events, featuring rich rhetorical expressions.
\end{itemize}

In the following tables, the values in parentheses denote the subset of texts identified as containing rhetorical devices (such as irony or sarcasm). These counts are derived from the model inference process validated in Section \ref{sec:appendix_rhetoric_validation}, highlighting the prevalence of pragmatic complexity in real-world scenarios.

\end{document}